# Unsupervised Real Time Prediction of Faults Using the Support Vector Machine

Dr Chen ZhiYuan
School of Computer Science
University of Nottingham, Malaysia Campus
Kuala Lumpur, Malaysia
Zhiyuan.Chen@nottingham.edu.my

Prof Dino Isa
Department of Electrical and Electronic Engineering
University of Nottingham, Malaysia Campus
Kuala Lumpur, Malaysia
Dino.Isa@nottingham.edu.my

*Abstract*— The research and development of Mission Critical System such as national communication system, national grid system and oil and gas pipeline network has been a great concern. Many innovative technologies have provided solutions and applications to the stability and continuous system operation. However a minute of interruption may cost millions of dollars and there were a lot of incident cases, which have brought bad impact to the economy. Therefore the purpose of this project is to design and implement a real time fault prediction system based on the support vector machine in order to meet the robustness requirements of a mission critical system. Current systems seldom implement the artificial intelligence such as Support Vector Machine due to the high cost of computational and no proven real-time artificial intelligence methods suitable for real-time systems. In this project we propose an intelligent system with an unsupervised real time prediction engine to detect faults. Meanwhile the issue of reducing the human intervention in the training phase has also been addressed by using an efficient strategic approach, which is a Zero Human Intervention Operation system that combines the support vector machine, clustering, and automatic parameter learning algorithms.

I. INTRODUCTION

Mission Critical Systems are used in many applications in Malaysia. The system usually addresses different types of applications such as those associated with communication systems, data storage systems, and power distribution systems. The systems need to operate continuously; a minute of interruption may cost millions of ringgits. For example on 30 April 2012, in Sabah 435,000 consumers were affected by the blackout for 10 hours due to a 66kV Capacitive Voltage Transformer explosion (New Strait Time Malaysia, 1 May 2012)[1]. As electricity is a necessity in all economic activity, the incident had a bad impact for the economy in Sabah on that day. On 20 April 2010, an oil spill severely affected the operation at BP-operated Macondo Prospect, and it has become known as the largest accidental marine oil spill in the history of the petroleum industry which has directly affected oil priced and subsequently affected world economy (USA Today, 2010)[2].

This has paved way for the development of fault finding systems using artificial intelligence. Current systems are unable to implement the computational artificial intelligence such as Support Vector Machine in real time because of the following limitations;

Firstly, the computational cost of artificial intelligence algorithm is expensive (M. Vogt and V. Kecman, 2005)[3]. In a mission critical system, robustness of every system is necessary; the complexity of current artificial intelligence algorithm may make the system vulbnerable to problems such as software crashes, overheating of the processor, and long execution time. Furthermore, the algorithm cannot be implemented in a mobile system becasue of the need for a sophisticated processing system (F.Oquendo, 2004)[4].

In this project, we address this issue by using efficient strategically approach for Quadratic Programming in order to solve Support Vector Machine objective function. The approach will maintain the Karush-Kuhn-Tucker (N. Christianini and J. Shawe-Taylor, 2000)[5] condition of the Quadratic Programming in order to obtain the correct result and good accuracy.( Z. Kolter and L. Honglak,2008)[6] This idea will reduce the computational cost of the algorithm to a certain level and meet the robustness requirements of a mission critical system.

Secondly, there is no proven artificial intelligence startegy for real-time systems. Real time systems require the AI method to process the acquired data one at a time. (Woodsend, K. 2009)[7] On the other hand, the artificial intelligence operation such as those associated with the support vector machine computes batch data in order to make decisions (A Shilton, 2006) [8]. In this project, we address this issue by using an incremental learning approach. As the data grows, the incremental learning approach usually requires a longer execution time; however, as discussed before, the proposed strategic approach of executing the Quadratic Programming will minimize this short coming. The incremental training will need to maintain the stability and the accuracy of the Support Vector Machine algorithm (K. Scheinberg, 2006)[9] to be better than or equal to batch processing.

Finally, conventional artificial intelligence systems need human intervention especially in the training phase. Mission critical system need to monitor continuously as the data will need to be collected over long periods of time. Thus, it is not practical for the system to require human for intervention. In this project, we address this issue by developing a Zero Human Intervention Operation system which combines the support vector machine, clustering, and automatic parameter learning algorithms. To implement the ideas generated for this project we intend to utilize a clustering technique based on the Self Organizing Map algorithm (J Vesanto, 1999) [10]

## II. RESEARCH APPROACH

### A. Development of Experimental Rig

An experimental rig is setup to emulate a Mission Critical System which is oil and gas pipeline system. The rig setup consists of four main components: pipeline with circulating oil, guided piezoelectric-ultrasonic transducer, a real time data acquisition and digital signal processing module.

Research Activity: To identify the effectiveness of various ultrasonic transducer arrangements, to determine appropriate mounting procedures and to experiment with the configuration of DAQ and DSP. The aim of this activity is to implement the most economical deployment of sensors without compromising accuracy and power usage.

### B. Data Processing Unit (Single Board Computer) and Wireless Transmission Setup

The single board computer will process data coming from the transducer. In addition, it controls the transducer parameters such as amplitude, frequency and waveform. Wireless transmission technology enables the transmission of processed data to the base station

Research Activity: To Identify the best processing cycle for Single Board Computer to minimize power consumption. To determine suitable wireless communication protocol in term of power consumption, efficiency, and speed. Experiment with different data compression technique to improve the data transmission and power efficiency.

### C. Design of Signal Conditioning Algorithm

This involves the design and implementation of a Digital Signal Processing Module to do suitable signal processing and conditioning such as filtering and compressing. The process cancels noise coming from the environment while retain significant signal for further classification and prediction.

Research Activity: To determine the best approach of signal conditioning while retaining the information signal from the transducer. Experiment with different filtering options such as the kalman-filter and moving average filter. The aim of this activity is to get a clean signal from the transducer in the presence of a noisy environment.

### D. Development of an Unsupervised Real-time Classification and Prediction Algorithm

This stage of the project requires investigations and development of a suitable algorithm that can process real-time data and make predictions. The technique must be developed for an unsupervised environment hence it must employ adaptive techniques in order to make it practical and robust. The aim of this stage is to produce an algorithm which can predict the time and amplitude of faults in a Mission Critical System as early as possible.

Research Activity: To manipulate the Quadratic Programming formulation in Support Vector Machine algorithm in order to transform it to a real time version. Current algorithm performs batch calculation of the training data whereas an online version requires the training to be done on a "data point by data point" basis.

## ACKNOWLEDGMENT

This work was supported by the Ministry of Science, Technology & Innovation, Malaysia. (Grant reference 01-01-12-SF0189).

## REFERENCES

[1] New Strait Time Malaysia, 1 May 2012
[2] USA Today, 20 April 2010
[3] M. Vogt and V. Kecman, "Active-Set Methods for Support Vector Machines," in Springer-Verlag book, 'Support Vector Machines: Theory and Applications', Ed. L. Wang, 2005, pp. 1-47.
[4] F.Oquendo, "π-ADL: an Architecture Description Language based on the higher-order typed π-calculus for specifying dynamic and mobile software architectures," ACM SIGSOFT Software Engineering Notes, , vol. 3, pp. 1-14, 2004.
[5] N. Christianini and J. Shawe-Taylor, "An Introduction to Support Vector Machines," Cambridge University Press, Cambridge, 2000.
[6] Z. Kolter and L. Honglak, "A Control Architecture for Quadruped Locomotion over Rough Terrain," in Proceedings of the IEEE International Conference on Robotics and Automation, 2008.
[7] K. Woodsend, "Hybrid MPI/OpenMP Parallel Linear Support Vector Machine Training," Journal of Machine Learning Research Vol.10 pp.1937-1953 2009 DBLP.
[8] A. Shilton, "Computational Intelligence Techniques", book chapter, Computational Intelligence for Movement Sciences: Neural Networks and other Emerging Techniques, Ed. Begg RK & Palaniswami M. IGI Publishing, USA. 396 pages, ISBN: 1-59140-836-9. 2006.
[9] K. Scheinberg, "An efficient implementation of an active-set method for SVMs," JMLR 7 (2006), pp. 2237-2257.
[10] J. Vesanto, "SOM-based data visualization methods," Intell. Data Anal., vol. 3, no. 2, pp. 111–126, 1999.